# A Better Match for Drivers and Riders: Reinforcement Learning at Lyft

Xabi Azagirre[a], Akshay Balwally[a], Guillaume Candeli[a], Nicholas Chamandy[a], Benjamin Han[a], Alona King[a], Hyungjun Lee[a], Martin Loncaric[a], Sébastien Martin[b*], Vijay Narasiman[a], Zhiwei (Tony) Qin[a], Baptiste Richard[a], Sara Smoot[a], Sean Taylor[a], Garrett van Ryzin[a], Di Wu[a], Fei Yu[a], Alex Zamoshchin[a]

[a] Lyft, San Francisco, California 94107 (current or formerly)
[b] Northwestern University, Evanston, Illinois 60208
* Corresponding author

**Contact**: xazagirre@lyft.com (XA), abalwally@lyft.com (AB), guillaumecandeli@gmail.com (GC), chamandy@waymo.com (NC), benjamin.han90@gmail.com (BH), alonak@lyft.com (AK), hyungjun23@gmail.com (HL), m.w.loncaric@gmail.com (ML), sebastien.martin@kellogg.northwestern.edu (SM), vnarasiman@lyft.com (VN), tqin@lyft.com (ZQ), sara.smoot@gmail.com(SS), brichard@lyft.com (BR), seanjtaylor@gmail.com (ST), garrett.vanryzin@gmail.com (GVR), wudimse@gmail.com (FY), azamoshchin@lyft.com (AZ)

## Abstract

To better match drivers to riders in our ridesharing application, we revised Lyft's core matching algorithm. We use a novel online reinforcement learning approach that estimates the future earnings of drivers in real time and use this information to find more efficient matches. This change was the first documented implementation of a ridesharing matching algorithm that can learn and improve in real time. We evaluated the new approach during weeks of switchback experimentation in most Lyft markets, and estimated how it benefited drivers, riders, and the platform. In particular, it enabled our drivers to serve millions of additional riders each year, leading to more than $30 million per year in incremental revenue. Lyft rolled out the algorithm globally in 2021.

# Introduction

Transportation is a trillion-dollar industry in the United States. After housing, Americans, on average, spend more of their income on transportation than on any other expense category, including food and healthcare. However, despite its importance to people's lives, transportation in North America today is fragmented and often inefficient. Personal automobiles account for approximately 98% of vehicle miles in the United States, despite being expensive and inconvenient in many contexts. Consequently, most North American cities have been built around cars rather than people. In several U.S. cities, 50% or more of the downtown area is dedicated to automobile infrastructure (Gardner 2011). Chester et al. (2010) estimate that there are a billion or more parking spaces in the United States—or approximately four spaces for each existing car. Meanwhile, the average personal vehicle sits idle about 95% of the time (Shoup 2011).

Lyft was created to improve people's lives by providing the world's best transportation. Its founders imagined a world built around people, with less pollution and traffic, where parks prevailed over parking lots, and where people spent less money and had more fun getting around. Achieving this mission required building a multimodal transportation network capable of reaching hundreds of millions of customers with diverse needs and constraints. Lyft's network has achieved impressive scale over the past decade, with growth opportunities still ahead—ridesharing accounts for only about 1.5% of total noncommercial vehicle miles traveled in the United States. Today, Lyft fulfills hundreds of millions of rider trips per year across a spectrum of transportation modes: ridesharing (private, on demand, or scheduled); traditional and electric bicycles; electric scooters; and public transit integrations. In doing so, Lyft provides earnings opportunities to millions of drivers and an array of personal transportation services to riders.

Lyft devotes many resources to building a compelling service with dependable mobile applications and an intuitive user interface, which combine to create seamless experiences for both riders and drivers. Nevertheless, growing Lyft's product portfolio and user base is just one aspect of the problem. Just as importantly, Lyft works to build and deploy optimization algorithms to increase the efficiency with which we move riders and drivers across the network every day. These include algorithms to allocate driver incentives and rider coupons efficiently, balance the marketplace in real time, route drivers to their destinations, swap electric vehicle batteries, redistribute bikes and scooters around a market, fight fraud, reduce insurance costs, and maximize the safety of our riders and drivers. The most fundamental algorithm of all—indeed, the first optimization algorithm developed in the early days of Lyft—is the real-time matching of drivers to riders.

Figure 1 presents a simplified overview of the primary algorithmic components of a rideshare network; see Wang and Yang (2019) for a comprehensive overview of ridesharing systems. It is no coincidence that the matching module is at the center of this diagram. A rider accesses the Lyft application with a transportation intent, possibly using a coupon, and checks the price and estimated time of arrival before making a ride request. (In practice, a rider might compare these parameters across a variety of travel options available in the Lyft application, such as luxury cars or bikes.) On the driver side of the marketplace, drivers receive financial incentives to become available on the Lyft platform and receive offers to fulfill specific rides. Vehicle routing and the allocation of driver repositioning incentives are other important algorithms that can improve throughput over the network. However, the matching algorithm is the most important one: it "clears" the marketplace by connecting the riders (demand) and drivers (supply) of the market in the most efficient manner possible. It must be extremely reliable because the matching system processes hundreds of millions of rider-driver matches every year.

**Figure 1.** The Graphic Illustrates the Interconnected Optimization Modules that Power a Ridesharing Network

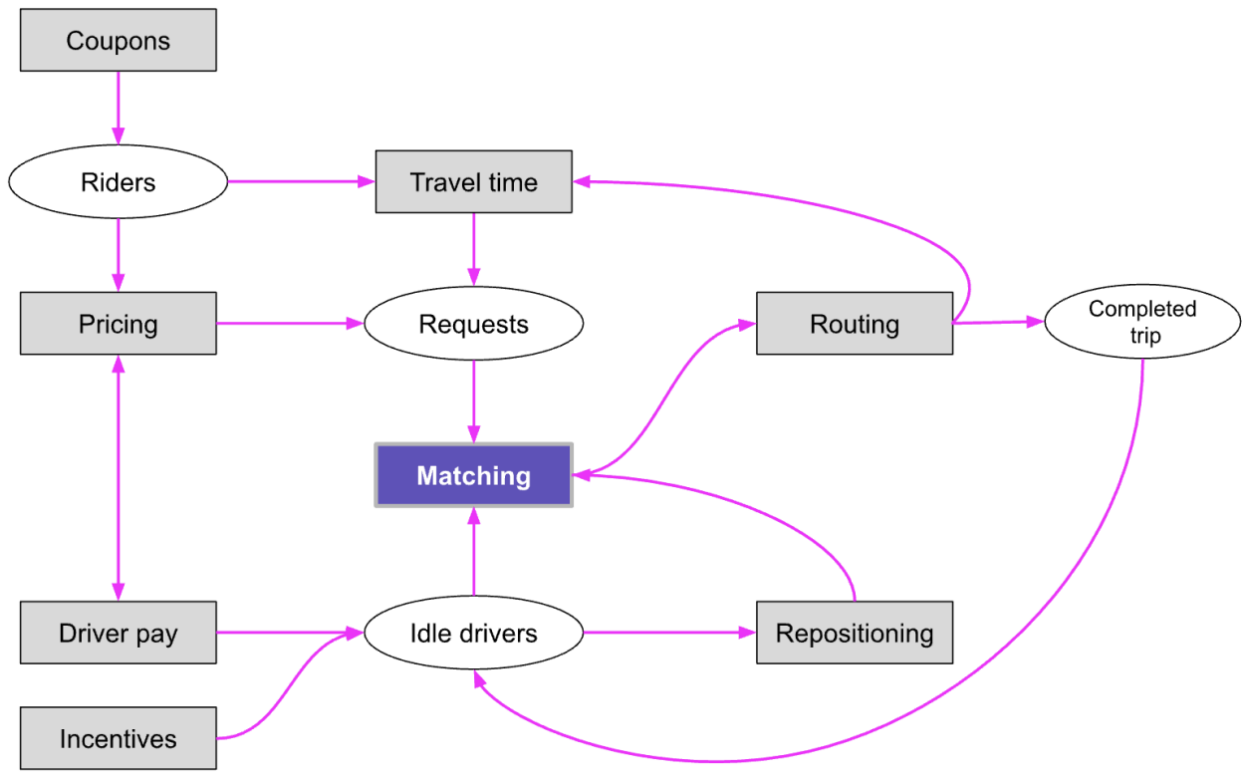

*Notes.* The arrows represent the information flows between the modules. Note, however, that the diagram may omit some modules and/or connections for simplicity.

In this paper, we describe how we completely redesigned Lyft's approach to matching. We implemented an online reinforcement learning (RL) approach. Although online RL is an active field of research with many success stories, it presents significant challenges. It is difficult to trust an algorithm that can change itself in real time, and examples of implementations in critical operational systems are rare. It is also hard to design an algorithm that can scale to the massive size of a ridesharing platform while still being fast enough to handle thousands of requests every minute. This may be why our work is the first documented real-world implementation of a ridesharing matching algorithm that can learn and improve in real time (Han et al. 2022). Nevertheless, we implemented this approach in all Lyft markets, enabling our drivers to serve millions of additional rides each year. Our extensive experiments (switchback) in most U.S. cities indicate that this change benefited all stakeholders: the drivers, the riders, and the platform.

## The Challenges of Online Matching

**The Online Matching Problem**

As rider requests arrive over time, the online matching problem aims to match these requests to drivers in the best possible way, to maximize metrics such as the total number of requests served (throughput) or the long-term driver earnings. When a request arrives, the platform can make one of the following decisions: match it with an available driver, delay the matching decision and wait for a better match, or reject the request. One could expect a simple strategy to work well: assigning the closest available driver to each rider, as Figure 2 shows. This strategy tries to greedily minimize the pickup time, which is intuitive: riders prefer a faster pickup and pickup time also reduces driver utilization and the platform throughput. However, this strategy can be far from optimal, especially if there are a few available drivers. In that case, the nearest driver may be far away, leading to a long pickup time. In turn, long pickup times mean that drivers must spend more time to serve each rider (i.e., pickup and serving time), decreasing the number of riders each driver can serve and further worsening the lack of available drivers. These downward spirals, which are referred to as the Wild Goose Chase phenomenon (Castillo et al. 2017), are risky for ridesharing platforms. A typical solution to this problem is to wait before assigning requests in an attempt to find better matches, or even reject requests. However, this requires access to precise demand forecasts.

Furthermore, matching a driver to a request often means that the driver will be available to serve another request at, or near, the destination of the first rider. The matching decisions, therefore, affect the distribution of available drivers in a city, with significant consequences to the number of requests the platform will be able to fulfill in the future. Good matching algorithms must try to anticipate the long-term consequences of the matches. For example, if a sufficient number of drivers is not available at an airport, matching requests going to the airport as quickly as possible may help mitigate the problem. In summary, designing an efficient and reliable online matching algorithm is a major challenge, which has been studied extensively at Lyft and in the scientific literature.

**Figure 2.** The Graphic Illustrates a Matching Decision: The Available Driver (1) Was Matched to a Rider and Will Drive to the Pickup Spot (Nearby Circle)

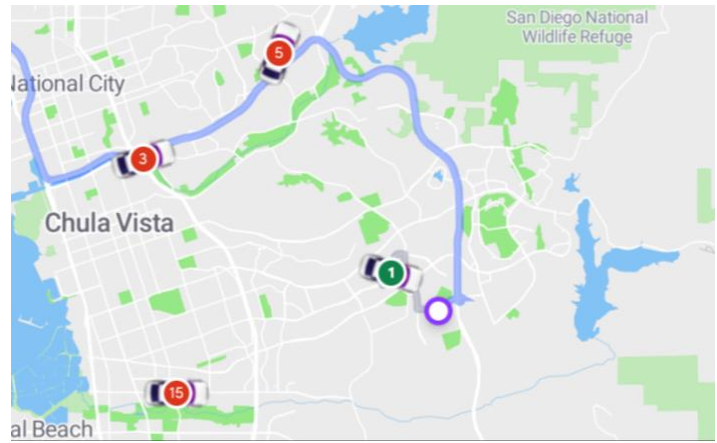

The problem of online matching that Lyft faced relates to the classical online bipartite matching problem in the literature. This is the optimization problem of maximizing the number of matches in a bipartite graph (e.g., the nodes on one side are the available drivers and the nodes on the other side are the available passengers) where nodes or edges appear dynamically. For example, Karp et al. (1990) introduce a worst-case optimal online policy in which the nodes on one side arrive one by one. Of course, these simplified models cannot accurately capture the complexities of real-world applications such as ours. Therefore, with the emergence of matching platforms, a large and recent body of literature adds more realistic features that are relevant to the ridesharing online matching problem. For example, Kanoria (2022) analyzes the asymptotic properties of a spatial online matching model, where the goal is to minimize the distance between the matched pairs (e.g., the total pickup time). Özkan and Ward (2020) also study the asymptotical properties of a queuing model of online matching and show that assigning the closest driver is often not optimal. Hu and Zhou (2022) consider the possibility of several types of supply and demand nodes (e.g., normal versus luxury drivers or patient versus impatient customers). Bertsimas et al. (2019) use New York City taxi data to show the benefits of using more complex optimization algorithms that take into account demand forecasts. Lowalekar et al. (2018) also consider future information using a multistage stochastic optimization approach. We refer to Yan et al. (2020) and Wang and Yang (2019) for a more thorough description of the rich ridesharing online matching literature.

Aouad and Saritaç (2020) show the limitation of "batching algorithms," which greedily match all available requests and drivers at regular time intervals. Although the above papers introduce model features that are relevant to Lyft's problem, they do not aim to solve the realistic optimization problems of ridesharing platforms, which include all of the above complexity and more. A particularly recent stream of literature also introduces the use of RL for realistic ridesharing platforms; see the review in Qin et al. (2022). However, these papers do not discuss the implementation of a fully online RL approach. We provide citations to the relevant RL papers as we introduce our approach in the *Online Reinforcement Learning* section.

**Batch Matching at Lyft**

Making a matching decision every time a new request comes would be impractical, given the high number of requests. Instead, every few seconds, Lyft uses *batch matching*. We gather all the unmatched requests and all the available drivers (i.e., a "batch") and determine which customer to match to which driver. This approach is illustrated on the right-hand side of Figure 3. Intuitively, batch matching can make smarter decisions than simple greedy approaches, which match customers one by one. For example, consider the simplistic scenario depicted on the left-hand side of Figure 3. Driver A is available and three minutes away from Rider 1 when Rider 1 requests a ride at 6 pm. At the same moment, Driver B is only two minutes away. Rider 2, requesting a ride momentarily at 6:01 pm, is 5 minutes and 10 minutes, respectively, from Drivers B and A. A greedy matching algorithm would assign Driver B to Rider 1, and Driver A to Rider 2. However, this is not the best match. Compared with the opposite assignment, it leads to a 50% increase (12 minutes versus 8 minutes) in the total time the riders spend waiting and the drivers spend navigating to the pickup locations. This example illustrates why accumulating a batch of ride requests before solving the assignment problem can be beneficial.

**Figure 3.** On the Left, the Graphic Illustrates a Stylized Example of a Ridesharing Matching Batch with Two Riders (1 and 2) and Two Available Drivers (A and B); on the Right, It Illustrates How Lyft Solves the Online Matching Problem as a Sequence of Batch Matching Decisions

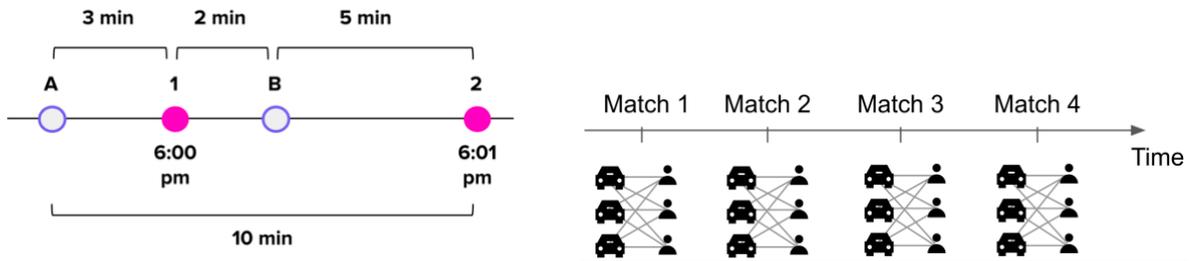

*Notes.* On the left, the minutes represent the pickup time between a rider and a driver. Before Rider 2 arrives, matching Driver B with Rider 1 seems optimal. However, when Rider 2 arrives, matching Rider 1 to Driver A and Rider 2 to Driver B is optimal.

Flexibility is an additional advantage of batch matching. To see this, consider the following mathematical formulation of batch-matching optimization:

$$\begin{aligned}
\max_{\mathbf{a} \in A} \quad & \sum_{i,j} w_{i,j}\, a_{i,j} \\
\text{s.t.} \quad & \sum_{i} a_{i,j} \leq 1 \quad \forall j \\
& \sum_{j} a_{i,j} \leq 1 \quad \forall i
\end{aligned} \quad (1)$$

In this formulation, the decision variables $a$ represent the matching decisions: $a_{i,j} = 1$ if we decide to match Rider $i$ and Driver $j$, and $a_{i,j} = 0$, otherwise. The two constraints correspond, respectively, to the fact that each Driver $j$ can be matched to at most one rider, and that each Rider $i$ can be matched to at most one driver (for simplicity we ignore shared rides here). Formally, the optimization formulation in Equation (1) corresponds to a maximum weight matching problem that Lyft must solve for each batch. Each potential match $(i, j)$ is associated with a weight $w_{i,j}$ representing how good the match is. The goal is to maximize the total weight (the objective function). This approach is particularly flexible, because Lyft can choose the vector of weights that best represents its objective (e.g., minimize pickup time, avoid wild goose chase). Every few seconds, Lyft uses an optimization method, for example, the Hungarian method (Kuhn 1955), to solve the corresponding massive bipartite matching problem in Equation (1). For tractability, not all rider-driver pairs need to be considered, and Lyft can filter these pairs by designing the feasible set $A$ in Equation (1) to limit the size of the problem. Most improvements to Lyft's matching algorithm over time can be mapped to a careful choice of weights and filters. Our novel RL approach also leverages this formulation, but automatically improves the choice of weights in real time, so that the corresponding matching decisions are best for the network in the long term.

**A Once-in-a-Century Shock to Transportation Systems**

We initiated our project during the COVID-19 pandemic. Despite having existed for only eight years, Lyft's business was subject to a once-in-a-century shock in early 2020 as the COVID-19 pandemic swept North America. Ridesharing demand decreased sharply for multiple reasons: companies transitioned to work-from-home models, bars and restaurants temporarily closed, and potential riders chose to reduce their risk of COVID-19 exposure. Nevertheless, demand did not drop to zero. Many of Lyft's users still relied on the platform, for example, when they needed to travel to work to do essential jobs or for transportation to medical appointments. Moreover, supply held up for a time, because drivers who rely on rideshare for their earnings tended to stay on the platform during the first weeks of the pandemic. Slowly, the marketplace began to settle into a new equilibrium, and with demand a fraction of what it had been in 2019, driver supply began to decrease to a commensurate level. The next two years saw a slow and steady return of rider demand to prepandemic levels. However, this recovery was not necessarily synchronized with the return of driver supply. It was also very heterogeneous across locations (i.e., at both regional and local levels), socioeconomic tiers, and other dimensions. Additionally, several mini shocks along the way (e.g., stimulus payments, new COVID-19 variants, gas price hikes) required constant readjustments of the marketplace equilibrium.

Lyft learned many lessons over this almost three-year period. From the perspective of using algorithms to control our marketplace, two learnings stood out early on: (1) Lyft's algorithms must be more resilient to marketplace shocks, large and small, and more adaptive to highly heterogeneous micro-conditions; and (2) Lyft's algorithms must function efficiently in a prolonged low-supply macro environment, which makes the matching problem particularly difficult. RL seemed like the perfect solution, and our project was born.

## Online Reinforcement Learning

### Reinforcement Learning in Ridesharing

RL is a machine learning paradigm we use to train a model through continuous interaction with the problem domain environment. The interaction is such that the agent (Lyft) executes an action (the matching decision) that affects the environment (e.g., the city, the drivers, the riders). In turn, the agent receives a feedback signal (e.g., a successful completion of a trip or the revenue from a trip) and can use this feedback to improve its decision making for the subsequent actions. Although RL has a reasonably long history, its use in solving complex large-scale decision problems has shown significant progress only over the past 10 years; see Silver et al. (2017) and Zheng et al. (2022).

Previously deployed works on RL for rideshare matching typically adopt an offline batch learning approach; see Qin et al. (2022) and the references therein, including Xu et al. (2018) and Qin et al. (2020). This approach trains the agent offline on a large set of static historical data, deploys the agent model online for real-time matching over an extended period while collecting interaction data at the data warehouse, and then iterates this process. In contrast, our work builds on an online RL framework, where the agent is updated in real time using the immediate feedback signals from the Lyft platform. This is the first full-scale industry deployment of an online RL method, to the best of our knowledge. We had to overcome many practical challenges; see the *Implementation and Impact* section. However, the result is a high-performing forward-looking rideshare matching algorithm that adapts well to a real-world environment.

The rideshare online matching problem can be modeled as a Markov decision process (MDP). Based on this process, we develop an online RL algorithm to learn and generate the optimal matching decisions in real time. There are multiple options for building the MDP model to solve this problem. The most direct approach is modeling the matching system as the *agent*. However, modeling the system as a single agent is challenging due to the combinatorial action space for batch assignment decisions. The approach that we take to break the curse of dimensionality is to decompose the system into the participating drivers (Xu et al. 2018).

By modeling the MDP around a driver, the *state s* of the agent is the driver's spatiotemporal position augmented by the supply-demand contextual features in the neighborhood. The agent performs an *action a* of either servicing a request by picking up the rider at the trip origin and transporting that rider to the destination or staying idle for a while. Because different actions generally span time intervals of different lengths, the MDP being developed here is, strictly speaking, a *semi-MDP* (Sutton et al. 1999). The *reward r* gained by performing an action is zero if the driver remains idle or becomes the driver's dollar earnings if the action is to match a request. Given the current state of the driver and a particular assignment action, we assume that the state transition is deterministic; that is, the next state of the driver is exactly the destination of the request or, in the case of idling, the following location. All the drivers in the system follow the same MDP model. Figure 4 shows an illustration of the state transitions in the MDP described.

**Figure 4.** The Graphic Illustrates the State Transitions for a Driver Agent

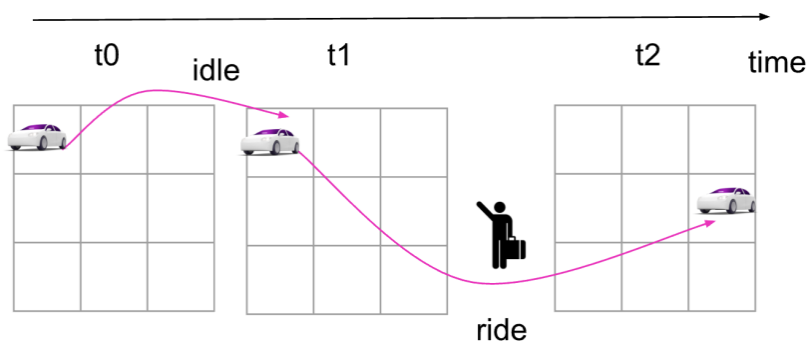

The *value function* $V(s)$ of the agent (driver) is the expected long-term discounted reward conditional on the current state, $V(s) := E$, where $\gamma$ is the time-discount factor. The action-value function $Q(s, a)$ is defined similarly, but conditional on both the current state and the action $a$. A policy $\pi(s): S \to A$ is a function that maps a state to an action. $S$ and $A$ are the state and action set, respectively. We provide an intuitive description of the driver value function in detail in the following paragraphs.

To better understand our online RL method, we briefly review *value iteration* (Sutton and Barto 2018), a dynamic programming method for solving an MDP through learning its optimal value function (and hence, optimal policy). Assuming that we know the state transition probabilities and the reward function, the value iteration method iteratively updates the current estimate of the optimal value function by following the Bellman optimality condition (Sutton and Barto 2018).

The vanilla value iteration algorithm (Sutton and Barto 2018) requires complete knowledge of the environment dynamics, which is unavailable in real-world rideshare scenarios. Another underlying assumption underpinning value iteration is stationarity in the environment. However, stationarity is also violated in the production environment of a rideshare platform, where the market condition (supply and demand) can be highly volatile, driven by nonrecurrent events and incidents. To ensure a high-performing approach that works robustly in practical settings, we developed an online approximate value iteration algorithm that learns and performs sample-based updates to the policy online, in contrast to an offline learning algorithm (see Qin et al. 2020) and on-policy (i.e., learning the value of the policy generating the data).

**A Driver's Value Function**

Our approach focuses on estimating the "value" of available drivers at any time and location in the city. This state value function $V(s)$ estimates how much money an available driver in a given state $s$ will expect to earn in the subsequent hours if that driver continues driving for Lyft, potentially serving many requests. We give the following example: A driver who logs onto the Lyft platform at 7 am in Oakland, California may secure a 40-minute ride to San Francisco and earn $45. Before logging off the Lyft platform, the driver may secure many other rides for a total of $300. Another driver, who starts at the same place at the same time, may not take as many trips or may decide to log off early, earns $150 in total. The driver value function of a driver in the state of (Oakland, 7 am) is the expected total earnings that an average driver at this location and time will make in the future–in this case $225 (($300+$150)/2).

Given the driver value function, we can evaluate a trip (or more generally, a transition), as Figure 4 shows, via advantage, a concept that quantifies the desirability of the trip in terms of the long-term value of the driver (and hence, the system). Intuitively, it is the expected extra earnings that a driver will make if we match the driver to a customer, including the potential long-term effects on the subsequent rides because the driver changed locations. It is the true "value" of a trip for the driver, not simply the immediate earnings of the trip. Formally, it is the difference between the action value $Q(s,a)$ and the state value $V(s)$, which tells us how different in value it is to perform a particular action (i.e., to match to a particular trip) at state $s$ than the average value of being at $s$. Mathematically, the advantage is

$$\Delta(s,a) = r + (1-p)(\gamma^d V(s') - V(s)),$$

where $s$ is the current driver state, $s'$ is the destination state, $r$ is the expected immediate reward (trip fare or 0 for idling), $d$ is the trip duration or four seconds for idling, $\gamma$ is the discount factor, and $p$ is the cancellation probability, which is 0 if idling.

The tabular form of the driver value function suffers from data sparsity. If there is no trip data involving a particular state $s$, then $V(s)$ will not be updated and hence may be poorly estimated. In our method, we address this problem by a particular value function approximation via linear factorization with coarse coding. Here, we represent each state $s$ by a set of relevant factors $k \in K(s)$. Examples of those factors are spatial partitions of geographical zones in geohash (a square grid system of spatial partitioning) (see Wikipedia 2023), time buckets, vehicle types, and their higher-order interaction terms. Then, the value function is approximated by

$$V(s) \approx \sum_{k \in K(s)} w_s(k) v(k),$$

where $v(k)$ is the factor value for $k$, and $w_s(k)$ is the corresponding weight, which is predetermined based on the relevancy of $k$ to $s$; for example, for geohash-based spatial factors, their weights are proportional to the inverse of the distance from the cell center to the driver's GPS coordinates. The factor values $v(k)$ are updated by the online value iterations, which we describe in the *Online Reinforcement Learning Framework* subsection below. Additional details of the value function approximation scheme can be found in Han et al. (2022). Deep learning-based approaches (e.g., Tang et al. 2019 and Holler et al. 2019) are also well-adapted to approximate the complex state space of the driver with high temporal and spatial granularity. However, this approach is not transparent enough to be used in real time for such a critical Lyft component.

In Figure 5, we can see the value of drivers in San Francisco during the night (left) and on a busy weekday morning (right). Higher values are in a darker shade. The pattern is clear and intuitive; at night, only the downtown area offers a high-earning prospect for the drivers. In contrast, on a weekday morning, the value for a driver is high over a much more expansive area around the Bay because of the commuting effect.

**Figure 5.** The Graphics Show a Heatmap Visualization of the Driver Value Function in San Francisco at Two Distinct Times (Left: Night; Right: Weekday Morning); Darker-Shade Areas Correspond to Higher Values

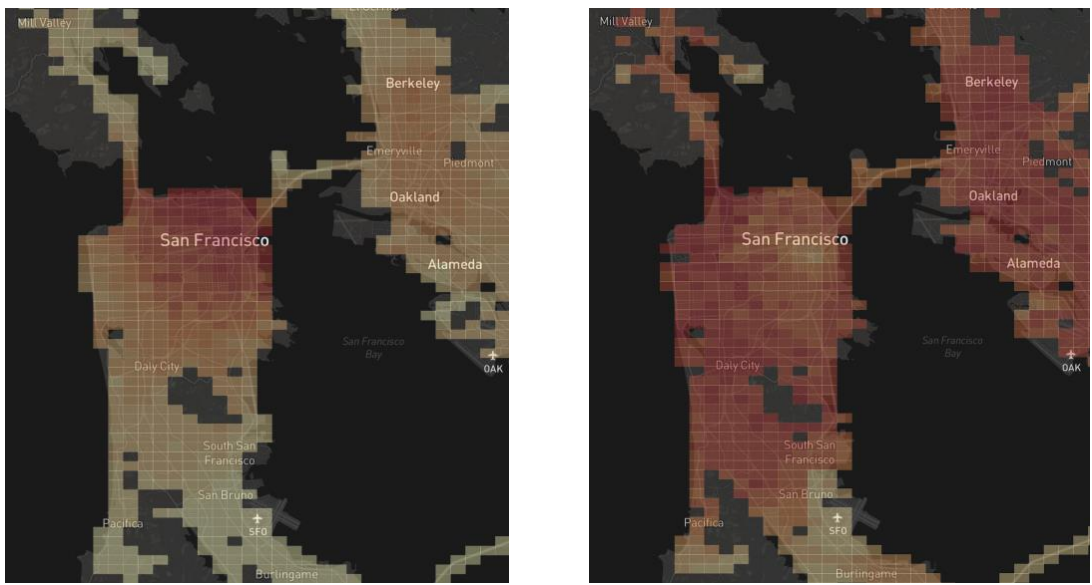

Now, we zoom into one San Francisco neighborhood and show in Figure 6 the normalized driver value function over a period of 14 days plotted by hours. We observe that the online value function captures three major aspects of the variations: (1) the daily periodicity with minima at late night and early morning, (2) the weekly periodicity, which typically peaks over the weekends, and (3) the week-over-week variations.

**Figure 6:** The Graph Illustrates the Time Series of the Driver Value Function for a San Francisco Neighborhood over Two Weeks

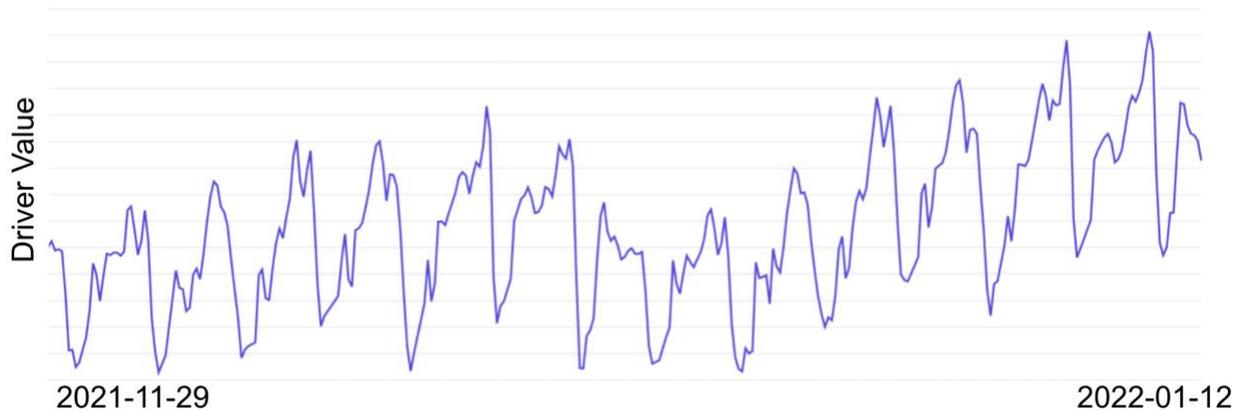

*Note.* The y-axis represents the normalized driver value.

The driver value function $V(s)$ lies at the core of Lyft's matching decision-making process, as we describe in the *Implementation and Impact* section below. The forward-looking nature of the state value function makes the matching decisions much less myopic and leads to an optimization over a longer horizon.

**Online Reinforcement Learning Framework**

The key novelty of our technical approach to the rideshare matching problem is an online RL framework that learns and updates the driver value function completely online and on-policy while generating the matching decisions in real time for our riders and drivers. In this subsection, we describe this framework in two parts: the system policy generation (improvement) and the online policy evaluation.

Recall, from *The Challenges of Online Matching* section and Equation (1), that our goal is to use the RL approach to find good matching weights $w_{i,j}$ for the batch matching problem system, which will determine how to match drivers to riders. Therefore, in RL terminology, selecting matching weights translates into a choice of policy. This step is illustrated on the right side of Figure 7. This is the policy improvement step in the value iterations algorithm that we introduced in the *Reinforcement Learning in Ridesharing* subsection. We compute $w_{i,j}$ to be the "value of the trip $i$ for driver $j$," as we describe in the subsection titled *A Driver's Value Function*. This is natural, because the value of the trip is by definition the extra earnings the driver will receive from this match: the matching problem simply tries to maximize the long-term driver earnings. Solving the batch-matching problem then approximates the maximization of the system value: we are improving our policy by leveraging the estimate of driver values. Nonetheless, these driver values are challenging to estimate because they are constantly changing based on the current market conditions and require planning several moves ahead. Our algorithm continually observes what is happening in the market and updates our estimates of driver value.

When the state transition probabilities are unknown, one typically resorts to sample-based updates, which are analogous to the stochastic gradient descent method used in deep learning. In our case, although the transition, given a driver state and a matching action, is deterministic, the demand arrivals are stochastic. Specifically, we use temporal difference (TD)-based value updates (Sutton and Barto 2018). The data involved in these updates are real-time rider requests (including trip fares), driver positions, and matching outcomes. For a matched request-driver pair, the corresponding trip fare serves as the immediate reward, and we compute the TD error by evaluating the driver value function at the driver's current position and the future destination. For an idle driver, the update is a decay in the value because the immediate reward is zero, and the transition is equivalent to a virtual trip of some standard length of time (the interval of a matching cycle). Using a tabular value function, the update is applied as follows:

$$V(s_i) \leftarrow V(s_i) + \alpha(r_{ij} + \gamma^{d_{ij}} V(s_{ij}) - V(s_i)),$$

where $s_i$ is the current driver state, $s_{ij}$ is the destination state or the successive idle driver state, $r_{ij}$ is the time-discounted trip fare or 0 for idle transitions, $d_{ij}$ is the trip duration or four seconds for idle transitions, and $\alpha$ is the learning rate. The update term (multiplied by the learning rate) is the TD error or the advantage associated with the transition. We refer the reader to Han et al. (2022) for further details on the updates under the value function approximation. The value function update corresponds to the policy evaluation step in the value iteration algorithm because it updates the value function to reflect the driver values under the new matching policy (see the bottom section of Figure 7).

Now, we are ready to describe the algorithmic framework of our online RL approach. The online approximate value iteration is a two-step iterative algorithm, which is analogous to a stochastic gradient descent method for solving large-scale machine learning problems. An iteration of the algorithm combines the steps discussed in this subsection:

1. Given the driver state value function (and hence, the approximate system value), compute the matching decisions for the drivers in the current cycle *n*.
2. Use the real-time dispatch feedback data to do one-step TD-updates to the driver value function.

**Figure 7.** The Graphic Depicts the Online RL (Approximate Value Iterations) Framework

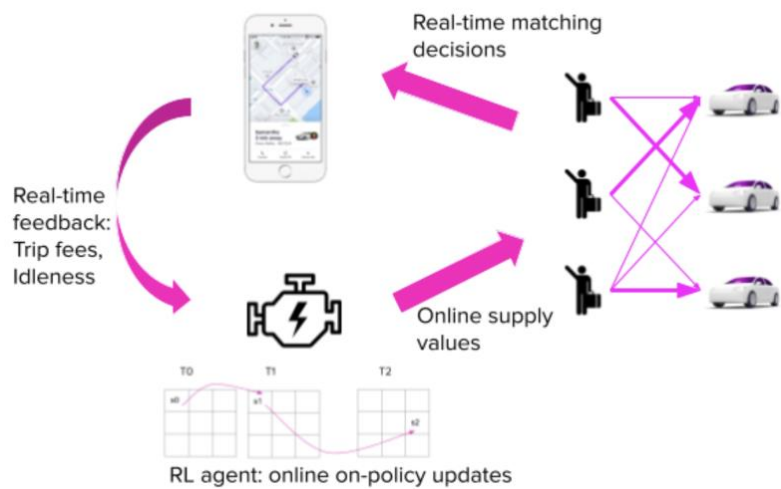

*Notes.* The online on-policy value updates represent the policy evaluation stage, corresponding to the bottom part of the figure. The real-time batch matching decisions represent the policy

improvement stage, which is illustrated by the matching graph and the arrow at the top-right-hand corner of the figure.

In summary, we use the values to make decisions, evaluate those decisions, and update them in real time to make even better decisions. Technically speaking, this novel approach corresponds to the online stochastic version of the value iteration algorithm. As we mention in the *Reinforcement Learning in Ridesharing* subsection, the original value iteration algorithm requires complete knowledge of the environment. Moreover, it inherently assumes that the environment is stationary. In a complex real-world setting like rideshare matching at Lyft, value iteration in its original form is unsuitable for practical implementation. Previously deployed solutions (e.g., Qin et al. 2020) adopt an offline RL approach, where policy evaluation is a computationally expensive offline step that learns on a large set of historical transition data. Once learned, the value function is deployed online for generating the matching policy and is not updated for an extended time (e.g., a week). Hence, it is an instance of the generalized policy iteration algorithm. Compared with previous offline RL-based solutions, our method allows value iteration to be implemented in an online and potentially nonstationary environment and has the crucial advantage of real-time updates and fast adaptivity to the rideshare environment variations. At the same time, it also encompasses innovative techniques to overcome the practical challenges to make it work in a real-world setting.

**Figure 8.** The Graph Provides a Comparison of the Driver Value Time Series for the Week of New Year's Eve (Undersupplied) in Miami and that of the Previous Week

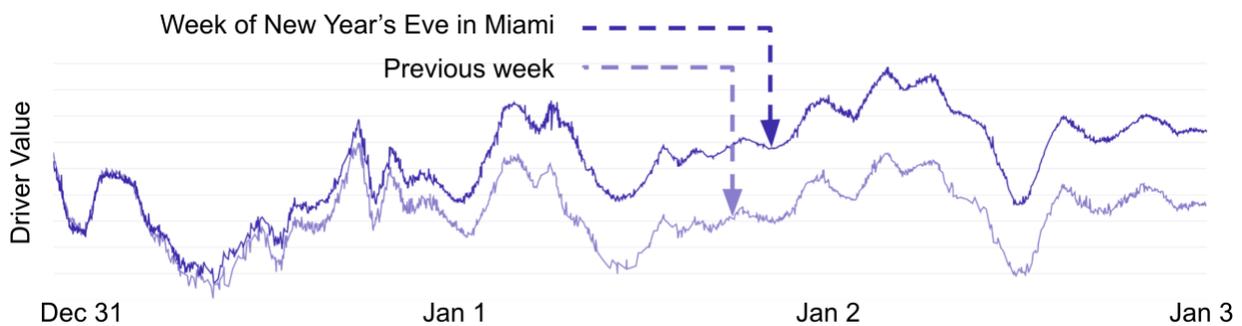

*Note.* The y-axis represents the normalized driver value.

We conclude this section with an illustration of how the online algorithm adapts to current conditions. We compare the driver value time series for the week of New Year's Eve in Miami with that of the previous week (see Figure 8). On New Year's Eve and the several days thereafter, there was both an increase in demand and lack of drivers, leading to an elevation of the driver value compared with the previous week. It is evident from the time series plot that the values at the beginning of the two series are similar. They start to diverge on New Year's Eve and converge again during the last two days of the week, which shows that our online learning algorithm for estimating the driver value reacts correctly to the change in the supply-demand condition. In the *Implementation and Impact* section below, we demonstrate through experimental results that our online RL method significantly improves the decision making for matching.

## Implementation and Impact

Moving from a research idea to an implementation is a long journey, especially for a component as critical as matching. We started to work on this project in early 2020, and we finally launched the change to all Lyft rideshare coverage areas in late 2021. In this section, we first briefly explain the various challenges of this implementation. We then discuss the culmination of our work: designing a massive month-long experiment in all Lyft regions to evaluate the algorithm. Finally, we use the experimentation results to quantify the estimated impact of the algorithm, which led to the final rollout decision.

### Implementation Challenges

Although RL is promising, it also presents many practical challenges, both technical and organizational. On the technical side, matching is at the core of Lyft's values, which means that our implementation must meet the highest reliability and stability bar, a significant software challenge. At peak load, our system must be able to support an extremely large number of driver value reads and updates per second. Additionally, we have greatly simplified the exposition for ease of understanding: a rider requests a ride; the rider is matched with a driver; and the driver picks up the rider. However, the reality rarely fits simple modeling assumptions, and implementation also requires extensive engineering work, practical RL expertise, realistic simulators, and parameter tuning.

As an example, we now list some features that our matching system must include but that we did not include in this paper's simplified model. First, either the driver or rider may initiate a cancellation. In addition, a driver who is matched to one rider can be "swapped" to match to another rider. Moreover, Lyft often matches a driver to a rider before the driver has dropped off a current rider. In addition, there are multiple types of rides in the Lyft marketplace. These include "standard" rides and rides in large and/or luxury vehicles. These types are not mutually exclusive, and drivers are often eligible to provide rides of multiple types. The concept of ride type must, therefore, enter the state space of the value function in a manner that is not combinatorially explosive with the number of types. These details may make the RL algorithm significantly more complex and/or create the need to update it regularly as the Lyft platform evolves. An initial version of our approach did not consider some subtle complexities of the marketplace, which led it to recommend suboptimal matching decisions in a nontrivial number of cases.

On the organizational side, a defining characteristic of online RL is that it interacts with and learns from any changes in its environment. However, implementing an RL matching algorithm could have unintended consequences in other areas of Lyft. Therefore, the alignment of many teams and stakeholders is necessary to launch such ambitious projects; however, this can be challenging in large companies with many products and teams. Lyft was no exception, as illustrated by the genesis of our project, which naturally emerged from conversations between people across separate teams and required us to move away from the organizational status quo.

Finally, the most challenging aspect of the project was evaluating it. Although simulators can give a sense of the performance of an algorithm, they are imperfect. It was imperative to try our approach in the real world and estimate its impact on the drivers, the riders, and the platform. Technology companies generally use experimentation approaches (such as A/B testing) to do so. However, we could not use A/B testing to test a change of the matching system. In *A Complex Experimental Design* below, we describe the costly experimentation process we had to follow, which involved constantly turning the RL algorithm on and off for an entire month across all locations where Lyft operates.

**A Complex Experimental Design**

We ran our tests in 2021 and structured them as time-split or "switchback" experiments. During "treatment" hours, but not during "control" hours, we augmented the matching algorithm with our RL approach. It is important to highlight the need for such an experimental design, because naively partitioning users into treatment and control groups would have biased the effect estimates due to the interaction between the two groups. Let us assume that half of our users are matched using our novel RL approach (treatment) and the other half using the myopic approach (control). By doing so, our RL approach would start to better balance the drivers throughout the city; however, this would affect both treatment and control users. That is, the treatment and control groups would not be independent, preventing us from accurately measuring the treatment effect. Thus, we measured matching algorithms at Lyft using a time-split experiment setup to enable (or disable) the new algorithm simultaneously for all the stakeholders (i.e., riders and drivers) in the marketplace.

Designing time-split experiments can be complex: we illustrate this with an example in Table 1. For this example, we cover three regions (e.g., major cities), and split each week into one-hour buckets (1 through 168). We start by randomly assigning each odd-hour bucket in Week 1 (shaded text) to the treatment or control group. We then assign the subsequent even bucket to the opposite variant. Given how consistent supply and demand patterns are across the week, we add a second week to the experiment to debias the estimator and assign each bucket to the opposite group of the one observed in Week 1. For example, if we randomly assign Bucket 1 to the control group in Week 1 (thus the word 'Random' in the brackets after 'Control' in Table 1), that will automatically assign Bucket 2 in Week 1 to the treatment group, and anti-symmetrically assign Bucket 1 to the treatment group (thus the word 'Paired' after 'Treatment') and Bucket 2 to the control group in Week 2. We run this process independently for each region. Using this design, we measure the average treatment effect of our success metrics by computing the difference between treatment and control sample hours. To avoid the transient effects between the buckets, we discard the data immediately before (burn-out period) and immediately after (burn-in period) each algorithm switch.

**Table 1.** The Table Provides a Simplified Example of Our Time-Split Experiment Setups

| Region, Hour | Week 1 Group | Week 2 Group | Region, Hour | Week 1 Group | Week 2 Group | Region, Hour | Week 1 Group | Week 2 Group |
|---|---|---|---|---|---|---|---|---|
| Region 1 Monday 12AM | Control (Random) | Treatment (Paired) | Region 2 Monday 12AM | Control (Random) | Treatment (Paired) | Region 3 Monday 12AM | Treatment (Random) | Control (Paired) |
| Region 1 Monday 1AM | Treatment (Paired) | Control | Region 2 Monday 1AM | Treatment (Paired) | Control | Region 3 Monday 1AM | Control (Paired) | Treatment |
| Region 1 Monday 2AM | Treatment (Random) | Control (Paired) | Region 2 Monday 2AM | Control (Random) | Treatment (Paired) | Region 3 Monday 2AM | Control (Random) | Treatment (Paired) |
| Region 1 Monday 3AM | Control (Paired) | Treatment | Region 2 Monday 3AM | Treatment (Paired) | Control | Region 3 Monday 3AM | Treatment (Paired) | Control |
| ... | ... | ... | ... | ... | ... | ... | ... | ... |
| Region 1 Sunday 10PM | Treatment (Random) | Control (Paired) | Region 2 Sunday 10PM | Treatment (Random) | Control (Paired) | Region 3 Sunday 10PM | Control (Random) | Treatment (Paired) |
| Region 1 Sunday 11PM | Control (Paired) | Treatment | Region 2 Sunday 11PM | Control (Paired) | Treatment | Region 3 Sunday 11PM | Treatment (Paired) | Control |

Although we expected the RL approach to better balance the drivers throughout the city, it required time before any benefits could be realized. Indeed, it could only assign drivers according to existing demand (e.g., by matching the drivers to specific requests). For example, even if our value function learns that the value of a driver in downtown New York City is high, we will not be able to increase supply there, unless there is demand heading in that direction. Therefore, our RL approach could require some time to significantly affect and improve the driver distribution in the city. In our experiments, we observed that after the RL policy was implemented (i.e., the experiment transitioned from control to treatment), most marketplace metrics required two to three hours to stabilize. This implied that if we wanted a steady state measure of the improved marketplace, we would need to set up our switchback experiment buckets to greater than four hours, which is an unusually long window at Lyft. Longer buckets implied fewer samples per week, which ultimately required significantly longer experimentation cycles to test any new iterations with sufficient statistical power. This is yet another example of the complexity of implementing RL in such a core system at Lyft: even validating its effectiveness proved to be challenging.

**A Successful Experiment**

We ran the switchback experiment in all the cities on our platform, divided into two main sets: (1) one for regions with high transaction volumes and high density, and (2) the second for more sparse and smaller regions; see Han et al. (2022) for details. Both sets of experiments showed that our new RL approach provided superior performance, improving outcomes for drivers, riders, and Lyft, a rare alignment in a two-sided marketplace. As we explain above in the *Online Reinforcement Learning* section, we could estimate the driver value function in real time, at any time and location, which allowed our matching algorithm to match the drivers and riders more efficiently. This improved the drivers' availability, resulting in shorter wait times and lower prices for riders. By doing so, during the experimentation period, we reduced the instances in which we could not find a driver to match for a ride request, which is arguably a rider's worst experience (see the explanation in Table 2), by 13%. In addition, rider cancellations decreased by 3% and riders' five-star ratings increased by 1%. These results collectively suggest that riders were experiencing better service levels and hence their satisfaction levels were higher. By doing so, the platform unlocked (1) higher earnings opportunities for drivers, and (2) millions of additional rides, which equates to over $30 million in annualized incremental revenue for Lyft, based on our internal estimation models. Moreover, these improvements were driven by algorithmic changes that required no budget allocation toward driver relocation.

**Table 2.** The Table Shows the Results We Achieved from Our Experiments Using the RL Approach

| Name | Description | Impact of RL Approach |
| --- | --- | --- |
| Unavailability | Ride requests for which we could not find a driver to match divided by total number of ride requests | -13.0% |
| Rider cancellation | Ride requests canceled by a rider divided by the total number of ride requests | -3.0% |
| Five-star ratings | Completed rides with five-star rating (maximum rating) divided by the total number of completed rides | +1.0% |
| Revenue (annualized) | Expected incremental revenue (with respect to the baseline) summed across the calendar year | >$30 million |

In summary, our experiments in both large and small regions were a resounding success; following the success of these experiments, Lyft rolled out our online RL approach across all its coverage areas in 2021.

**Optimization for Rideshare and Trade-offs**

When iterating on one of the many optimization algorithms that jointly managed the Lyft marketplace, there was a trade-off: Lyft could benefit in one dimension but be disadvantaged in another. The challenge was to tune our algorithm so that such a trade-off was worthwhile. Managing trade-offs is a reality of optimizing Lyft's network: trade-offs between different business metrics, ride requests competing for scarce supply, improving both the riders' and the drivers' experiences, and short-term and long-term objectives. For example, as a market-balancing mechanism, Lyft might apply a temporary price increase during a period of extreme undersupply. This could have the benefit of making Lyft's service more reliable and shortening the pickup time when a rider and a driver are matched. However, it has a downside in that it makes rideshare temporarily less affordable and may cause some price-sensitive users to avoid riding with Lyft in the future or lose them to competition. The matching problem illustrates this paradigm. For example, as we saw in the simple example in Figure 3, using longer windows for solving the matching graph can greatly improve efficiency. By reducing pickup times, the Lyft platform can fulfill more rides in the same amount of time and pass on the savings to users, both riders and drivers, through lower prices or higher earnings. However, a longer matching window means that a rider must spend more time waiting to find out which driver has been assigned and when that driver will arrive. This can make the Lyft experience feel slower, create anxiety for the rider, or cause the rider to consider alternative options.

Although trade-offs in metrics are the norm in rideshare optimization, there are a few rare exceptions where algorithmic innovation improves the entire Pareto frontier. The experimental results in Table 2 show how our project falls into that category. Improving the matching algorithm is extremely hard because the status quo was already the outcome of years of continuous experimentation and refinement. Therefore, the success of our project is a testimony to the strength of online RL.

## Conclusions

This project demonstrated the viability of an industry-scale, adaptive learning system applied to a fundamental problem underpinning modern transportation. Applications of online RL to critical operational systems are still rare, because trusting an algorithm that can update itself is difficult for many people. In this paper, we discussed RL and its significant implementation challenges. However, we proved that overcoming these obstacles is worthwhile: we significantly improved on a status quo that had been perfected over many years. This improvement benefited riders, drivers, and the platform, an advantage that is extremely rare for a ridesharing platform. Although RL adds significant complexity, we are now convinced that its benefits outweigh its challenges.

We believe that this work is also transportable to other settings. At Lyft, we continue to iterate on our matching algorithm and work on extensions of this approach to other core components of the marketplace. More broadly, the general online matching problem is not limited to ridesharing. Logistics systems, like food or package delivery, and inventory management systems share similar characteristics. However, at this time, online RL use is mostly limited to games, simulations, or relatively low-risk applications. Indeed, as we demonstrated in this paper, there are many challenges to implementing complex RL approaches in practice, and it is hard to trust an algorithm that can change itself. Nonetheless, we hope our application can encourage other actors to use RL for other critical operational decisions.